%% file: acl_latex.tex
\pdfoutput=1

\documentclass[11pt]{article}

\usepackage[final]{acl}

\usepackage{times}
\usepackage{latexsym}

\usepackage{microtype}
\usepackage{hyperref}
\usepackage{url}
\usepackage{booktabs}

\usepackage{natbib}
\usepackage{graphicx}
\usepackage{adjustbox}
\usepackage{amsmath}
\usepackage{amsfonts}
\usepackage{caption}
\usepackage{subcaption}
\usepackage{wrapfig}
\usepackage[most]{tcolorbox}
\usepackage{balance}

\usepackage[T1]{fontenc}    

\usepackage[utf8]{inputenc}

\usepackage{microtype}

\usepackage{inconsolata}

\usepackage{graphicx}

\newcommand{\para}[1]{\vspace{0.5em}\noindent\textbf{#1}}
\newcommand{\edc}{{{\textsc{EDC}}}}
\newcommand{\edcr}{{{\textsc{EDC+R}}}}
\newcommand{\fon}[1]{\fontfamily{#1}\selectfont}

\newcommand{\edcta}{{{Target Alignment}}}
\newcommand{\edcsc}{{{Self Canonicalization}}}
\newcommand{\edcsr}{{{Schema Retriever}}}

\title{Extract, Define, Canonicalize: An LLM-based Framework for Knowledge Graph Construction
}

\author{
 \textbf{Bowen Zhang\textsuperscript{1}} and
 \textbf{Harold Soh\textsuperscript{1,2}}
\\
\\
 \textsuperscript{1}Dept. of Computer Science, National University of Singapore,
 \textsuperscript{2}NUS Smart Systems Institute
\\
 \small{
   \href{mailto:{bowenzhang, harold}@comp.nus.edu.sg}{\{bowenzhang, harold\}@comp.nus.edu.sg}
 }
}

\begin{document}
\maketitle
\begin{abstract}
In this work, we are interested in automated methods for knowledge graph creation (KGC) from input text. Progress on large language models (LLMs) has prompted a series of recent works applying them to KGC, e.g., via zero/few-shot prompting. Despite successes on small domain-specific datasets, these models face difficulties scaling up to text common in many real-world applications. A principal issue is that, in prior methods, the KG schema has to be included in the LLM prompt to generate valid triplets; larger and more complex schemas  easily exceed the LLMs' context window length. Furthermore, there are scenarios where a fixed pre-defined schema is not available and we would like the method to construct a high-quality KG with a succinct self-generated schema. To address these problems, we propose a three-phase framework named Extract-Define-Canonicalize (\edc{}): open information extraction followed by schema definition and post-hoc canonicalization. \edc{} is flexible in that it can be applied to settings where a pre-defined target schema is available and when it is not; in the latter case, it constructs a schema automatically and applies self-canonicalization. To further improve performance, we introduce a trained component that retrieves schema elements relevant to the input text; this improves the LLMs' extraction performance in a retrieval-augmented generation-like manner. 
We demonstrate on three KGC benchmarks that \edc{} is able to extract high-quality triplets without any parameter tuning and with significantly larger schemas compared to prior works. Code for EDC is available at \url{https://github.com/clear-nus/edc}.
\end{abstract}

\input{content/intro}

\input{content/background}
\input{content/method}

\input{content/experiments}

\input{content/discussion}

\input{content/limitations}
\input{content/ethics}
\input{content/acks}

\balance
\bibliography{custom}

\appendix

\nobalance

\input{content/appendix}

\end{document}

%% file: content/intro.tex
\section{Introduction}

\begin{figure}[t!]
    \centering
    \includegraphics[width=0.5\textwidth]{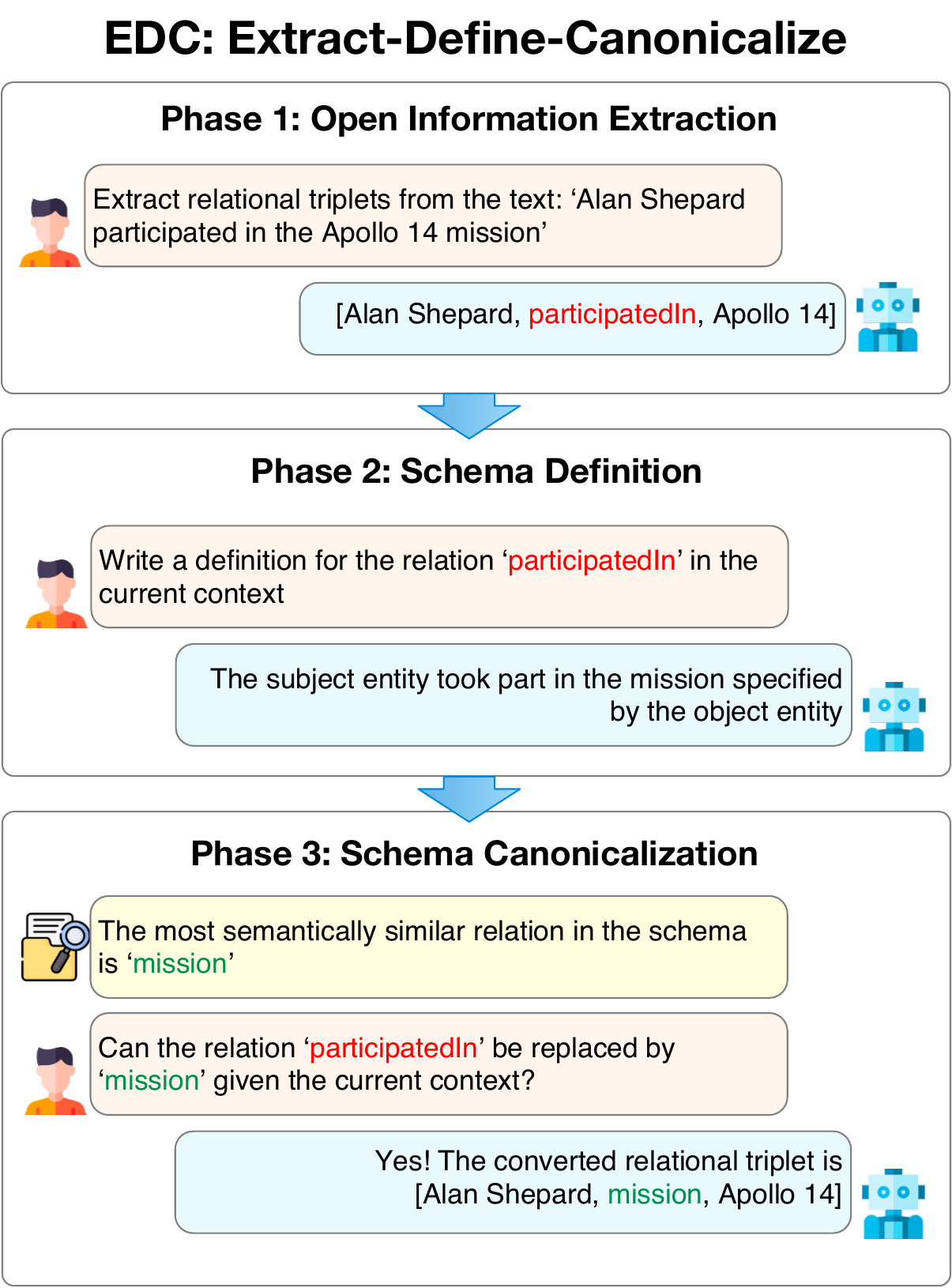}
    \caption{A high-level illustration of Extract-Define-Canonicalize (EDC) for Knowledge Graph Construction.}
    \label{fig:intro_fig}
\end{figure}

Knowledge graphs (KGs)~\citep{ji2021kgsurvey} are a structured representation of knowledge that organizes interconnected information through graph structures, where entities and relations are represented as nodes and edges. They are broadly used in a variety of downstream tasks such as decision-making~\citep{guo2021kgfordeicsionmaking, lan2020kgfordecisionmaking}, question-answering~\citep{huang2019kgforqa, yasunaga2021kgforqa}, and recommendation~\citep{guo2020kgforrs, wang2019kgforrs}. However, knowledge graph construction (KGC) is inherently challenging: the task requires competence in understanding syntax and semantics to generate a consistent, concise, and meaningful knowledge graph. As such, KGC predominantly relies on intensive human  labor~\citep{ye2022generativekgcsurvey}. KGC is a broad problem and in this work, \textbf{we focus on the task of relational triplet extraction} as it is crucial for KGC. Following previous works~\citep{ye2022generativekgcsurvey, melnyk2022grapher, bi2024codekgc}, we still refer to the task we are addressing as KGC.

Recent attempts to automate KGC~\citep{zhong2023automatickgcsurvey, ye2022generativekgcsurvey} have employed large language models (LLMs) in view of their remarkable natural language understanding and generation capabilities. LLM-based KGC methods employ various innovative prompt-based techniques, such as multi-turn conversation~\citep{wei2023chatie} and code generation~\citep{bi2024codekgc}, to generate entity-relation triplets that represent the knowledge graph. However, these methods are currently limited to small and domain-specific scenarios --- to ensure the validity of generated triplets, schema information (e.g., possible entity and relation types) has to be included in the prompt. Complex datasets (e.g., Wikipedia) typically require \textbf{large schemas that exceed the context window length} or can be ignored by the LLMs~\citep{wadhwa2023revisiting}. Furthermore, \textbf{pre-defined schemas are not always available} --- the users might not have pre-determined or fixed intentions about what information is of interest in advance but still would like to extract intrinsically high-quality KGs. It is unclear how existing methods will work in such situations.

To address these problems, we propose \textbf{Extract}-\textbf{Define}-\textbf{Canonicalize} (\textbf{\edc{}}), a structured approach for KGC: the key idea is to decompose KGC into three primary phases corresponding to three subtasks (Fig. \ref{fig:intro_fig}):

\begin{enumerate}
\item Open Information Extraction: extract a list of entity-relation triplets from the input text freely. 
\item Schema Definition:  generate a definition for each component of the schema, e.g. entity type and relation type, induced by triplets obtained in the extraction phase. 
\item Schema Canonicalization: use the schema definitions to standardize the triplets such that semantically-equivalent entities/relations types have the same noun/relation phrase.
\end{enumerate}

Each phase exploits the strengths of LLMs: the Extract subtask leverages recent findings that LLMs are effective open information extractors~\citep{li2023evaluatingchatgptie, han2023isiesolvedbychatgpt} --- they can extract semantically correct and meaningful triplets. However, the resulting triplets typically contain redundant and ambiguous information, e.g., multiple semantically equivalent relation phrases such as `profession', `job', and `occupation'~\citep{kamp2023oieandcanonsurvey, putri2019aligning, vashishth2018cesi}. 

Phases 2 and 3 (Define and Canonicalize) standardize the triplets to make them useful for downstream tasks. We designed \edc{} to be flexible: it can either discover triplets consistent with a pre-existing schema of potentially large size (\textbf{\edcta}) or \textit{self-generate} a schema (\textbf{\edcsc}). To achieve this, we use LLMs to define the schema components by exploiting their explanation generation capabilities --- LLMs can justify their extractions via explanations that are agreeable to human experts~\citep{li2023evaluatingchatgptie}. The definitions are used to find the closest entity/relation type candidates (via a vector similarity search) that the LLM can then reference to canonicalize a component. In the case there is no equivalent counterpart in the existing schema, we can choose to add it to enrich the schema.

To further improve performance, the three steps above can be followed by an additional \textbf{Refinement} phase: we repeat EDC but provide the previously extracted triplets and a relevant part of the schema in the prompt during the initial extraction. We propose a trained \textbf{\edcsr{}} that retrieves schema components relevant to the input text, akin to retrieval-augmented generation~\citep{lewis2020rag}, which we find improves the generated triplets.  

Experiments on three KGC datasets in both \edcta{} and \edcsc{} settings show that \edc{} is able to extract higher-quality KGs compared to state-of-the-art methods through both automatic and manual evaluation. Furthermore, the use of the \edcsr{} is shown to significantly and consistently improve \edc{}'s performance.

In summary, the paper makes the following contributions:

\begin{itemize}
    \item \edc{}, a flexible and performant LLM-based framework for knowledge graph construction that is able to extract high-quality KGs with schema of large size or without any pre-defined schema.
    \item \edcsr{}, a trained model to extract schema components relevant to input text in the same vein as information retrieval.
    \item Empirical evidence that demonstrate the effectiveness of \edc{} and the \edcsr{}.
\end{itemize}

%% file: content/background.tex
\section{Background}
In this section, we provide relevant background on knowledge graph construction (KGC), open information extraction (OIE), and canonicalization. 

\paragraph{Knowledge Graph Construction.}
Traditional methods typically addressed KGC using ``pipelines'', comprising subtasks like entity discovery~\citep{vzukov2018entitydiscovery, martins2019entitydiscovery}, entity typing \citep{choi2018entitytyping, onoe2020entitytyping}, and relation classification \citep{zeng2014relationclassification, zeng2015relationclassification}. Thanks to advances in pre-trained generative language models (e.g.,  T5~\citep{raffel2020t5} and BERT\citep{lewis2019bart}), more recent works instead frame KGC as a sequence-to-sequence problem and generate relational triplets in an end-to-end manner by fine-tuning these moderately-sized language models~\citep{ye2022generativekgcsurvey}. The success of large language models (LLMs) has pushed this paradigm further: current methods directly prompt the LLMs to generate triplets in a zero/few-shot manner. For example, ChatIE~\citep{wei2023chatie} extracts triplets by framing the task as a multi-turn question-answering problem and CodeKGC~\citep{bi2024codekgc} approaches the task as a code generation problem. As previously mentioned, these models face difficulties scaling up to general text common in many real-world applications as the KG schema has to be included in the LLM prompt. Our \edc{} framework circumvents this problem by using post-hoc canonicalization (and without requiring fine-tuning of the base LLMs).

\paragraph{Open Information Extraction and Canonicalization.}
Standard (closed) information extraction requires the output triplets to follow a pre-defined schema, e.g. a list of relation or entity types to be extracted from. In contrast, open information extraction (OIE) does not have such a requirement. OIE has a long history and we refer readers who want comprehensive coverage to the excellent surveys~\citep{liu2022oiesurvey, zhou2022neuraloiesurvey, kamp2023oieandcanonsurvey}. Recent studies have found LLMs to exhibit excellent performance on OIE tasks~\citep{li2023evaluatingchatgptie}. However, the relational triplets extracted from OIE systems are not canonicalized;  multiple semantically equivalent relations can coexist without being unified to a canonical form, causing redundancy and ambiguity in the induced open knowledge graph. An extra canonicalization step is required to standardize the triplets to make the KGs useful for downstream applications. 

Canonicalization methods differ depending on whether a target schema is available. In case a target schema is present, the task is sometimes referred to as ``alignment''~\citep{putri2019aligning}. For example, \citet{putri2019aligning} uses WordNet~\citep{miller1995wordnet} as side information to obtain definitions for the OIE-extracted relation phrases and a Siamese network to compare an OIE relation definition and a pre-defined relation in the target schema. In case no target schema is available, state-of-the-art methods are commonly based on clustering~\citep{vashishth2018cesi, dash2020vaecanon}. CESI~\citep{vashishth2018cesi} creates embeddings for the OIE relations using side information from external sources like PPDB~\citep{ganitkevitch2013ppdb} and WordNet. However, clustering-based methods are prone to over-generalization~\citep{kamp2023oieandcanonsurvey, putri2019aligning}, e.g., CESI may put ``is brother of'', ``is son of'', ``is main villain of'', and ``was professor of'' into the same relation cluster.

Compared to the existing canonicalization methods, \edc{} is more general; it works whether a target schema is provided or not. Instead of using static external sources like WordNet, \edc{} utilizes contextual and semantically-rich side information generated by LLMs. Furthermore, by allowing the LLMs to verify if a transformation can be performed (instead of solely relying on the embedding similarity), EDC alleviates the over-generalization issue faced by previous methods.

%% file: content/method.tex
\section{Method: EDC for KGC}
\label{sec: method}
This section outlines our primary contribution: an approach to constructing knowledge graphs that leverages LLMs in a structured manner. We first detail the \edc{} framework followed by a description of refinement (\textbf{\edcr}). Given input text, our goal is to extract relational triplets in a canonical form such that the resulting KGs will have minimal ambiguity and redundancy. When there is a pre-defined target schema, all generated triplets should conform to it. In the scenario where there is not one, the system should dynamically create one and canonicalize the triplets with respect to it. 

\subsection{EDC: Extract-Define-Canonicalize}
At a high level, EDC decomposes KGC into three connected subtasks. To ground our discussion, we will use a specific input text example: ``\textit{Alan Shepard was born on Nov 18, 1923 and selected by NASA in 1959. He was a member of the Apollo 14 crew}'' and walk through each of the phases:

\para{Phase 1: Open Information Extraction:} we first leverage Large Language Models (LLMs) for open information extraction. Through few-shot prompting, LLMs identify and extract relational triplets ([Subject, Relation, Object]) from input texts, independent of any specific schema. Using our example above, the prompt is:
\begin{tcolorbox}[breakable, fontupper=\fon{cmss}, title=OIE Prompt]
\begin{footnotesize}
Given a piece of text, extract relational triplets in the form of [Subject, Relation, Object] from it.

Here are some examples:

Example 1:

Text: The 17068.8 millimeter long ALCO RS-3 has a diesel-electric transmission.

Triplets: [[`ALCO RS-3', `powerType', `Diesel-electric transmission'], [`ALCO RS-3', `length', `17068.8 (millimetres)']]
...

Now please extract triplets from the following text:
Alan Shepard was born on Nov 18, 1923 and selected by NASA in 1959. He was a member of the Apollo 14 crew.
\end{footnotesize}
\end{tcolorbox}

The resultant triplets (in this case, \textsf{[`Alan Shepard', `bornOn', `Nov 18, 1923']}, \textsf{
[`Alan Shepard', `participatedIn', `Apollo 14’]}) form an \textit{open KG}, which is forwarded to subsequent phases. 

\para{Phase 2: Schema Definition:} Next, we prompt the LLMs to provide a natural language definition for each component of the schema induced by the open KG:
\begin{tcolorbox}[fontupper=\fon{cmss}, breakable, enhanced jigsaw, title=Schema Definition Prompt]
{\begin{footnotesize}
Given a piece of text and a list of relational triplets extracted from it, write a definition for each relation present.

Example 1:

Text: The 17068.8 millimeter long ALCO RS-3 has a diesel-electric transmission.

Triplets: [[`ALCO RS-3', `powerType', `Diesel-electric transmission'], [`ALCO RS-3', `length', `17068.8 (millimetres)']]

Definitions:

powerType: The subject entity uses the type of power or energy source specified by the object entity.

... 

Now write a definition for each relation present in the triplets extracted from the following text:

Text: Alan Shepard was an American who was born on Nov 18, 1923 in New Hampshire, was selected by NASA in 
1959, was a member of the Apollo 14 crew and died in California

Triplets: [[`Alan Shepard', `bornOn', ‘Nov 18, 1923’],  [`Alan Shepard', `participatedIn', `Apollo 14’]]

\end{footnotesize}}
\end{tcolorbox}
This example prompt results in the definitions for
(\textsf{bornOn: The subject entity was born on the date specified by the object entity.})
and (\textsf{participatedIn: The subject entity took part in the event or mission specified by the object entity.}), 
which are then passed to the next stage as \textit{side information} used for canonicalization.

\para{Phase 3: Schema Canonicalization:} The third phase aims to refine the open KG into a canonical form, eliminating redundancies and ambiguities. We start by vectorizing the definitions of each schema component using a sentence transformer to create embeddings. Canonicalization then proceeds in one of two ways, depending on the availability of a target schema:

\begin{itemize}

\item \edcta{}: With an existing target schema, we identify the most closely related components within the target schema for each element, considering them for canonicalization. To prevent issues of over-generalization, LLMs assess the feasibility of each potential transformation. If a transformation is deemed unreasonable, indicating no semantic equivalent in the target schema, the component, and its related triplets are excluded.

\item \edcsc{}: Absent a target schema, the goal is to consolidate semantically similar schema components, standardizing them to a singular representation to streamline the KG. Starting with an empty canonical schema, we examine the open KG triplets, searching for potential consolidation candidates through vector similarity and LLM verification. Unlike target alignment, components deemed non-transformable are added to the canonical schema, thereby expanding it.

\end{itemize}
Using our example, the prompt is: 
\begin{tcolorbox}[breakable, fontupper=\fon{cmss}, enhanced jigsaw, title=Schema Canonicalization Prompt]
\begin{footnotesize}

Given a piece of text, a relational triplet extracted from it, and the definition of the relation in it, choose the most appropriate relation to replace it in this context if there is any.

Text: Alan Shepard was born on Nov 18, 1923 and selected by NASA in 1959. He was a member of the Apollo 14 crew.

Triplets: [`Alan Shepard', `participatedIn', `Apollo 14']

Definition of `participatedIn': The subject entity took part in the event or mission specified by the object entity.

Choices:

A. `mission': The subject entity participated in the event or operation specified by the object entity.

B. `season': The subject entity participated in the season of a series specified by the object entity.

...

F. None of the above
\end{footnotesize}
\end{tcolorbox}
Note that the choices above are obtained by using vector similarity search. After the LLM makes its choice, the relations are transformed to yield: \textsf{[`Alan Shepard’, `birthDate', `Nov 18, 1923']}, \textsf{
[`Alan Shepard’, `mission', `Apollo 14']}, which forms our canonicalized KG.

\subsection{EDC+R: iteratively refine EDC with Schema Retriever}

The refinement process leverages the data generated by EDC to enhance the quality of the extracted triplets. Inspired by retrieval-augmented generation and prior work~\cite{bi2024codekgc}, we construct a ``hint'' for the extraction phase (details in Appendix~\ref{appendix: refinement_hint_detail}), which comprises two main elements: 
\begin{itemize}
    \item Candidate Entities:  The entities extracted by \edc{} from the previous iteration, and entities extracted from the text using the LLM;
    \item Candidate Relations: The relations extracted by \edc{} from the previous cycle and relations retrieved from the pre-defined/canonicalized schema by using a trained \edcsr{}. 
\end{itemize}
The inclusion of entities and relations from both the LLM and the schema retriever provides a richer pool of candidates for the LLM, which addresses issues where the absence of entities or relations impairs the LLM's effectiveness. By merging the entities and relations extracted in earlier phases with new findings from entity extraction and schema retrieval, the hint serves to aid the OIE by bootstrapping from the previous round. 

To scale EDC to large schemas, we employ a trained \edcsr{}  which allows us to efficiently search schemas. The Schema Retriever works in a similar fashion to information retrieval methods based on vector spaces~\cite{ganguly2015informationretrieval, lewis2020rag}; it projects the schema components and the input text to a vector space such that cosine similarity captures the relevance between the two, i.e., how likely a schema component to be present in the input text. Note that in our setting, the similarity space is different from the standard sentence embedding models where cosine similarity in the vector space captures semantic equivalence. Our Schema Retriever is a fine-tuned variant of the sentence embedding model E5-mistral-7b-instruct~\citep{wang2023e5mistral}. We follow the original training methodology detailed in the paper, which involves utilizing pairs of text and their corresponding defined relations. For details, please refer to the Appendix~\ref{appendix: schema_retriever_training}. For a given positive text-relation pair $(t^+,r^+)$, we employ an instruction template on $t^+$ to generate a new text $t^+_{inst} =$ ``Instruct: retrieve relations that are present in the given text $\textbackslash n$ Query: $\{t^+\}$''.

We then finetune the embedding model to distinguish between the correct relation associated with a given text and other non-relevant relations using the InfoNCE loss.

Back to our example, refinement with the schema retriever adds the following relation to the previous set: \textsf{[`Alan Shepard', `selectedByNasa', `1959']}. The relation \textsf{`selectedByNasa'} is rather obscure but was specified in the target schema.

%% file: content/experiments.tex
\section{Experiments}

In this section, we describe experiments designed to evaluate the performance of \edc{} and \edcr{}. Briefly, our results demonstrate that \edc{} significantly outperforms the state-of-the-art methods in both \edcta{} and \edcsc{} settings. Refinement further improves \edc{}. Source code for EDC and to replicate our experiments are available in the supplementary materials, with full tables in the Appendix~\ref{appendix: target_alignment_results}.

\subsection{Experimental Setup}

\paragraph{Datasets.}
We evaluate \edc{} using three KGC datasets:
\begin{itemize}
    \item WebNLG~\citep{ferreira2020webnlg}: We use the test split from the semantic parsing task of WebNLG+2020 (v3.0). It contains 1165 pairs of text and triplets. The schema derived from these reference triplets encompasses 159 unique relation types.
    \item REBEL~\citep{cabot2021rebel}: The original test partition of REBEL comprises 105,516 entries. To manage costs, we select a random sample of 1000 text-triplet pairs. This subset induces a schema with 200 distinct relation types.
    \item Wiki-NRE~\citep{distiawan2019wikinre}: From Wiki-NRE's test split (29,619 entries), we sample 1000 text-triplet pairs, resulting in a schema with 45 unique relation types.
\end{itemize}

These datasets were chosen due to their richer variety of relation types over alternatives like ADE~\citep{gurulingappa2012ade} (1 relation type), SciERC~\citep{luan2018scierc} (7 relation types), and CoNLL04~\citep{roth2004conll} (4 relation types) used to evaluate previous LLM-based methods~\citep{bi2024codekgc, wadhwa2023revisiting}. This diversity better mimics real-world complexity. In our experiments, we focus on extracting relations as the only schema component available across all datasets. Relations, being a foundational element of KGs, are prioritized over other components like entity or event types. However, note that \edc{} can be readily extended to other schema components.

\paragraph{EDC Models.}

\edc{} contains multiple modules that are powered by LLMs. Since the OIE module is the key upstream module that determines the semantic content captured in the KG, we tested different LLMs of different sizes including GPT-4~\citep{achiam2023gpt4}, GPT-3.5-turbo~\citep{brown2020gpt3}, and Mistral-7b~\citep{jiang2023mistral}. Mistral-7b was deployed on a local workstation, whereas the GPT models were accessed via the OpenAI API. For the framework's remaining components which required prompting, we used GPT-3.5-turbo. In the canonicalization phase, the E5-Mistral-7b model was utilized for vector similarity searches without modifications.

\begin{figure*}[h!]
    \begin{subfigure}[t]{.33\textwidth}
        \includegraphics[width=1\textwidth]{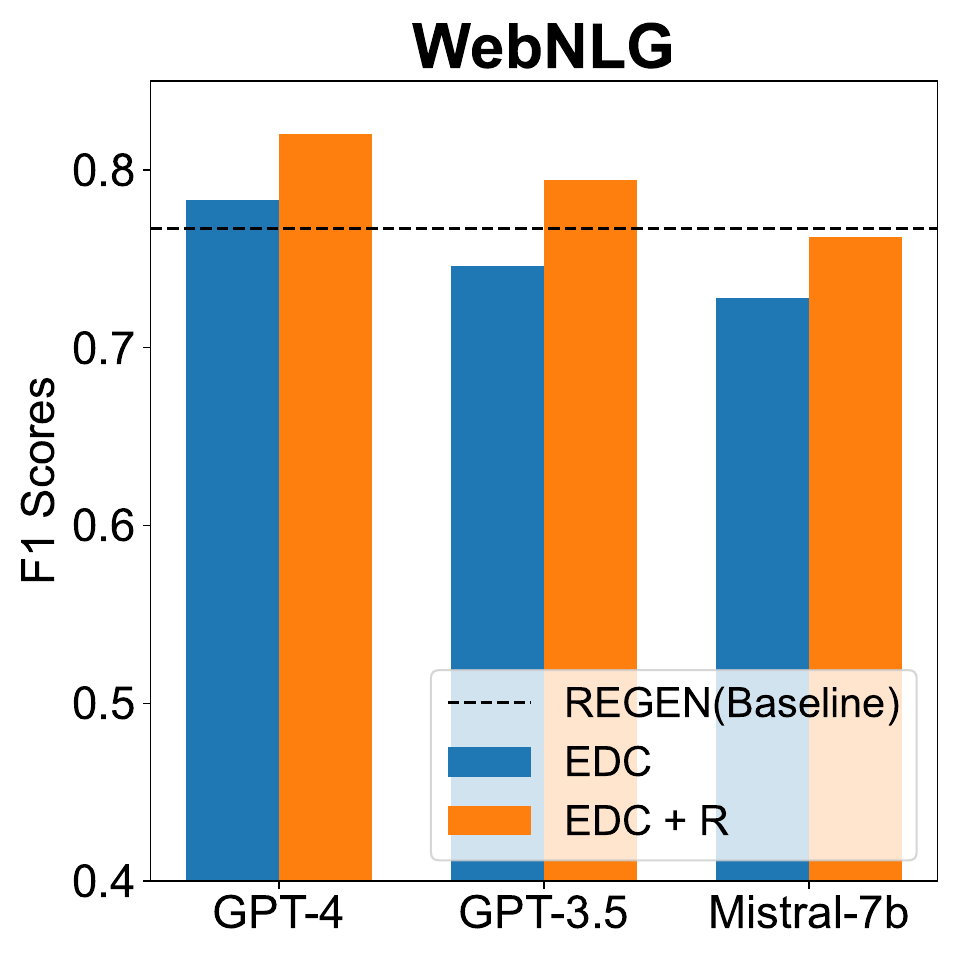}
        \label{fig:webnlg_barchart}
    \end{subfigure}
    \begin{subfigure}[t]{.33\textwidth}        \includegraphics[width=1\textwidth]{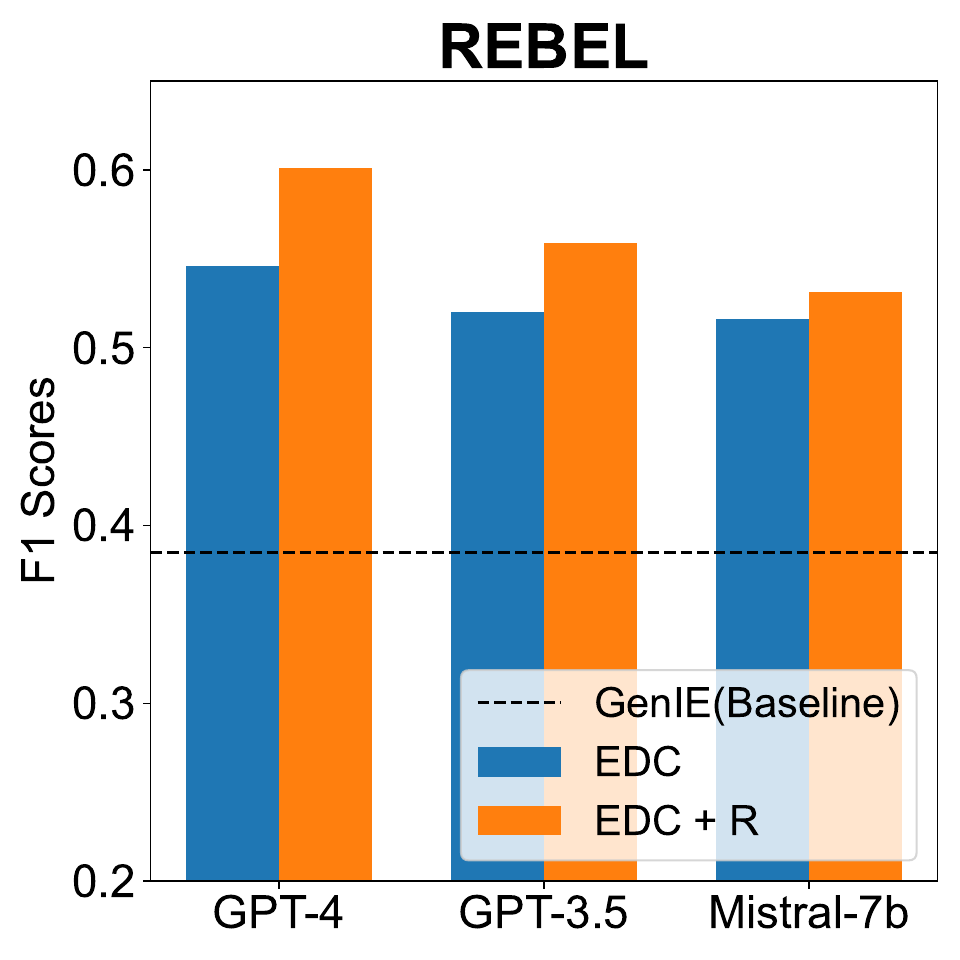}
        \label{fig:rebel_barchart}
    \end{subfigure}
    \begin{subfigure}[t]{.33\textwidth}
        \includegraphics[width=1\textwidth]{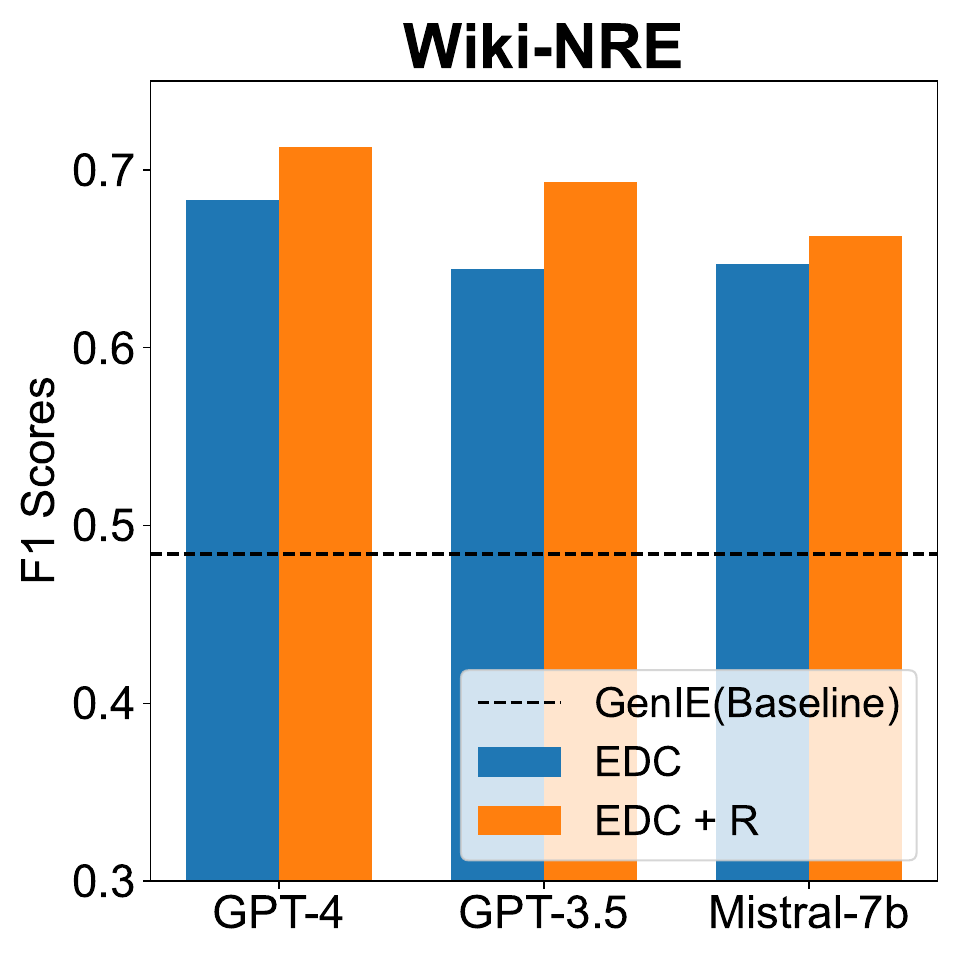}
        \label{fig:wikinre_barchart}
    \end{subfigure}
    \caption{Performance of \edc{} and \edcr{} on WebNLG, REBEL, and Wiki-NRE datasets against baselines in the \edcta{} setting (F1 scores with `Partial' criteria). \edcr{} only performs one iteration of refinement due to diminishing marginal improvement.}
    \label{fig:ta_barchart}
    \vspace{-10pt}
\end{figure*}

\subsubsection{Evaluation Criteria and Baselines}

We evaluate our methods differently under Target Alignment (when a schema is provided) and Self Canonicalization (no schema) due to the \textit{inherently different objectives}: the former aims to recover the ground-truth annotated triplets consistent with the target schema while the latter is to extract semantically correct and meaningful triplets that induce a succinct and non-redundant KG without a pre-defined target to compare against. For the datasets above, the preivous LLM-based KGC methods (ChatIE and CodeKGC) could not be used due to the schema size. Although EDC is not intended for small domain-specific datasets, we include the results on SciERC and CoNLL04 in the Appendix~\ref{appendix: small_datasets_results} for the comprehensiveness of the evaluation.

\paragraph{Target Alignment.} We compare \edc{} and \edcr{} against the specialized trained models for each of the datasets:
\begin{itemize}
    \item \textbf{REGEN}~\citep{dognin2021regen} is the SOTA model for WebNLG. It is a sequence-to-sequence model that leverages pre-trained T5~\citep{raffel2020t5} and Reinforcement Learning (RL) for bidirectional text-to-graph and graph-to-text generation.  
    \item \textbf{GenIE}~\citep{josifoski-etal-2022-genie}, a sequence-to-sequence model that leverages pre-trained BART~\citep{lewis2019bart} and a constrained generation strategy to constrain the output triplets to be consistent with the pre-defined schema. GenIE is the state-of-the-art model for REBEL and Wiki-NRE.
\end{itemize}
Following previous work~\citep{dognin2021regen, melnyk2022grapher}, we use the WEBNLG evaluation script~\citep{ferreira2020webnlg} which computes the Precision, Recall, and F1 scores for the output triplets against the ground truth in a token-based manner. Metrics based on Named Entity Evaluation were used to measure the Precision, Recall, and F1 score in three different ways.
\begin{itemize}
\item \textit{Exact:} Requires a complete match between the candidate and reference triple, disregarding the type (subject, relation, object).
\item \textit{Partial:} Allows for at least a partial match between the candidate and reference triple, disregarding the type.
\item \textit{Strict:} Demands an exact match between the candidate and reference triplet, including the element types.
\end{itemize}

\paragraph{\edcsc{}.} For evaluating  self-canonicalization performance, comparisons are made with:
\begin{itemize}
    \item \textbf{Baseline Open KG}, which is the initial open KG output from the OIE (Open Information Extraction) phase. This serves as a reference point to illustrate the changes in precision and schema conciseness resulting from the canonicalization process.
    \item \textbf{CESI}~\citep{vashishth2018cesi}, recognized as a leading clustering-based approach for open KG canonicalization. By applying CESI to the open KG, we aim to contrast its performance against canonicalization by \edc.
\end{itemize}
Given that canonicalized triplets may use relations phrased differently from the reference triplets or entirely out-of-schema relations, a token-based evaluation becomes unsuitable. Thus, we resort to manual evaluation, focusing on three key aspects that reflect the intrinsic quality of an extracted KG:
\begin{itemize}
    \item \textit{Precision:} The canonicalized triplets remain correct and meaningful with respect to the text compared to the OIE triplets.
    \item \textit{Conciseness:} The schema's brevity is measured by the number of relations types.
    \item \textit{Redundancy:} We employ a redundancy score --- the average cosine similarity among each canonicalized relation and its nearest counterpart --- where low scores indicate that the schema's relations are semantically distinct.
\end{itemize}

\subsection{Results and Analysis}
\label{result_subsection}
In the following, we focus on conveying our main findings and results. For full results and tables, please refer to the Appendix. 

\subsubsection{Target Alignment}
\label{target_alignment}

The bar charts in Figure~\ref{fig:ta_barchart} summarize the Partial F1 scores obtained by \edc{} and \edcr{} on all three datasets with different LLMs for OIE compared against the respective baselines. \textbf{\edc{} demonstrates performance that is superior to or on par with the state-of-the-art baselines for all evaluated datasets}. Comparing the LLMs, GPT-4 emerges as the top performer, with Mistral-7b and GPT-3.5-turbo exhibiting comparable results. The disparity between our methods and the baselines is more pronounced on the REBEL and Wiki-NRE datasets; this is primarily due to the GenIE's constrained generation approach, which falls short in extracting triplets that include literals, such as numbers and dates.

\textbf{Refinement (\edcr{}) consistently and significantly enhances performance}. Post-refinement, the difference in performance between GPT-3.5-turbo and Mistral-7b is larger, suggesting Mistral-7b's was not as able to leverage the provided hints. Nevertheless, a single refinement iteration with the hint improved performance for all the tested LLMs.

From the scores, it appears that \edc{} performance is significantly better on WebNLG compared to REBEL and Wiki-NRE. However, we observed that \edc{} was penalized despite producing valid triplets on the latter datasets. A reason for this is that the reference triplets in these datasets are non-exhaustive. For example, given the text in the REBEL dataset, `\textit{Romany Love is a 1931 British musical film directed by Fred Paul and starring Esmond Knight, Florence McHugh and Roy Travers.}', \edc{} extracts: \textsf{[`Romany Love', `cast member', `Esmond Knight']}, \textsf{[`Romany Love', `cast member', `Florence McHugh']}, \textsf{[`Romany Love', `cast member', `Roy Travers']}, which are all semantically correct, but only the first triplet is present in the reference set. The datasets also contain reference triplets based on information extraneous to the text, e.g., `\textit{Daniel is an Ethiopian footballer, who currently plays for Hawassa City S.C.}' has a corresponding reference triplet \textsf{[`Hawassa City S.C.', `country', `Ethiopia']}. %

These issues can be attributed to the distinct methodologies employed in the creation of these datasets. For WebNLG, annotators were asked to compose text solely from the triplets. Thus, the text and the triplets have a direct correspondence, and the text typically does not include information other than what is apparent from the triplets. In contrast, REBEL and Wiki-NRE are created by aligning text and triplets using distant supervision~\citep{smirnova2018distantsupervision}. This approach can result in less straightforward triplet extraction and incomplete reference sets, leading to overly pessimistic evaluations for methods like \edc{}, which generate correct triplets not present in the dataset.~\citep{han2023isiesolvedbychatgpt, wadhwa2023revisiting}. 
On average, EDC extracts one additional triplet per sentence on REBEL and Wiki-NRE compared to the reference set, while on WebNLG, it extracts a similar number of triplets to the reference.

\input{content/tbl_retriever_ablation_vertical}

\para{Ablation study on schema retriever.} 
To evaluate the impact of the relations provided by the schema retriever during refinement, we conducted an ablation study with GPT-3.5-turbo by removing these relations. The results in Table~\ref{tab: tbl_sr_ablation_vertical} show that   \textbf{ablating the \edcsr{} leads to a  decline in performance}. Qualitatively, we find that the schema retriever helps to find relevant relations that are challenging for the LLMs to identify during the OIE stage. For example, given the text \textit{`The University of Burgundy in Dijon has 16,800 undergraduate students'}, the LLMs extract \textsf{[`University of Burgundy', `location', `Dijon']} during OIE. Although semantically correct, this relation overlooks the more specific relation present in the target schema, namely \textsf{`campus'}, for denoting university's location. The schema retriever successfully identifies this finer relation, enabling the LLMs to adjust their extraction to \textsf{[`University of Burgundy', `campus', `Dijon']}. This experiment highlights the schema retriever's value in facilitating the extraction of precise and contextually appropriate relations.

\subsubsection{Self Canonicalization}

\input{content/tbl_canon_vertical}

Here, we focus on evaluating \edc's self-canonicalization performance (utilizing GPT-3.5-turbo for OIE). We omit refinement in \edcsc{} setting as it has already been studied above and in subsequent iterations, the self-constructed canonicalized schema becomes the target schema. %
Following prior work~\citep{wadhwa2023revisiting, kolluru2020openie6}, we conducted a targeted human evaluation of knowledge graphs. This evaluation involved two independent annotators assessing the reasonableness of triplet extractions from given text without prior knowledge of the system's details. We observed a high inter-annotator agreement score of 0.94.

The evaluation results and schema metrics are summarized in Table~\ref{tab: sc_results_vertical}.While the open KG generated by the OIE stage contains semantically valid triplets (which affirms the previous findings that LLMs are competent open information extractors~\cite{li2023evaluatingchatgptie}), there is a significant degree of redundancy within the resultant schema. \textbf{\edc{} accurately canonicalizes the open KG and yields a schema that is both more concise and less redundant compared to CESI}. EDC avoids CESI's tendency toward over-generalization --- in line with prior work~\citep{putri2019aligning}, we observed CESI inappropriately clusters diverse relations such as \textsf{`place of death'}, \textsf{`place of birth'}, \textsf{`date of death'}, \textsf{`date of birth'}, and \textsf{`cause of death'} into a single \textsf{`date of death'} category.

%% file: content/tbl_retriever_ablation_vertical.tex
\begin{table}[]\centering
\caption{Ablation study results (F1 scores with all criteria) on schema retriever, the LLM used for OIE is GPT-3.5-turbo. S.R. stands for \edcsr{}.}\label{tab: tbl_sr_ablation_vertical}
\scriptsize
\begin{tabular}{l|l|ccc}\hline\hline
Dataset& Method &Partial &Strict &Exact \\\hline
&EDC+R &\textbf{0.794} &\textbf{0.753} &\textbf{0.772} \\
WebNLG &EDC+R w/o S.R. &0.752 &0.701 &0.721 \\\cline{2-5}
&EDC &0.746 &0.688 &0.713 \\\hline
&EDC+R &\textbf{0.559} &\textbf{0.516} &\textbf{0.529} \\
REBEL &EDC+R w/o S.R. &0.517 &0.466 &0.482 \\\cline{2-5}
&EDC &0.506 &0.449 &0.473 \\\hline
&EDC+R &\textbf{0.693} &\textbf{0.685} &\textbf{0.657} \\
Wiki-NRE &EDC+R w/o S.R. &0.653 &0.645 &0.641 \\\cline{2-5}
&EDC &0.647 &0.638 &0.640 \\
\hline\hline
\end{tabular}
\end{table}

%% file: content/tbl_canon_vertical.tex
\begin{table}[]\centering
\caption{Performance of EDC in the Self Canonicalization setting (human-evaluated precision and schema metrics). The best result for each dataset and metric is bolded. Prec. stands for precision, No. Rel. stands for the number of relations and Red. stands for redundancy score.}\label{tab: sc_results_vertical}
\scriptsize
\begin{tabular}{l|l|ccc}\hline\hline
Dataset&Method &Prec.$(\uparrow)$ &No. Rel.$(\downarrow)$ &Red.$(\downarrow)$ \\\hline
&EDC &\textbf{0.956} &\textbf{200} &\textbf{0.833} \\
WebNLG &CESI &0.724 &280 &0.893 \\\cline{2-5}
&Open KG &0.982 &529 &0.927 \\\hline
&EDC &\textbf{0.867} &\textbf{225} &\textbf{0.831} \\
REBEL &CESI &0.504 &307 &0.854 \\\cline{2-5}
&Open KG &0.903 &667 &0.895 \\\hline
&EDC &\textbf{0.898} &\textbf{106} &\textbf{0.833} \\
Wiki-NRE &CESI &0.753 &114 &0.849 \\\cline{2-5}
&Open KG &0.909 &204 &0.881 \\
\hline\hline
\end{tabular}
\end{table}

%% file: content/discussion.tex
\section{Conclusion}

In this work, we presented \edc{}, an LLM-based three-phase framework that addresses the problem of KGC by open information extraction followed by post-hoc canonicalization. Experiments show that \edc{} and \edcr{} are able to extract better KGs than specialized trained models when a target schema is available and dynamically create a schema when none is provided. The scalability and versatility of \edc{} opens up many opportunities for applications: it allows us to automatically extract high-quality KGs from general text using large schemas like Wikidata~\citep{vrandevcic2014wikidata} and even enrich these schemas with newly discovered relations.

%% file: content/limitations.tex
\section{Limitations and Future Directions}

There are several limitations that we would like to address in future works. 

\begin{itemize}
    \item We only considered schema canonicalization within the scope of this paper, it is of great interest to incorporate an entity de-duplication mechanism in the future to reduce the redundancy in the constructed KGs, e.g., via coreference resolution~\citep{sukthanker2020corefsurvey}. We briefly explored this approach and the preliminary results can be found in  Appendix~\ref{appendix: edc_with_other_tools}.
    \item EDC's components can be further improved to boost performance. Specifically, the schema retriever may benefit from training on more diverse and higher-quality data. 
    \item Due to time and resource constraints, we only tested different LLMs for OIE while all the other modules of EDC rely on GPT-3.5-turbo, it will be beneficial to test the smaller open-source models' performance on the other tasks as well.
    \item EDC is a costly framework, involving a large number of LLM calls. When GPT-3.5-turbo is used for all components, the cost was around 0.009 USD per example in our experiments. It is possible to have certain components replaced by smaller fine-tuned models --- previous works have shown smaller language models can be fine-tuned for OIE~\citep{wadhwa2023revisiting} and smaller BERT-based classifiers can be trained for schema canonicalization. We also explored the possibility of combining the two stages of OIE and Schema Definition in  Appendix~\ref{combining_oie_and_sd}.
    \item We are looking to apply EDC towards embodied AI and robotics. Specifically, KGs can form memory sources for VLMs, containing facts about humans~\cite{zhang2023large}, the task or goal~\cite{xie2023translating}, and the environment. 
\end{itemize}

%% file: content/ethics.tex
\section{Ethical Considerations}
\paragraph{Artifact usage.} The datasets we used in the paper are only leveraged for research purposes and we strictly follow the corresponding licenses (e.g. WebNLG uses cc-by-nc-sa-4.0). It is to be noted that, due to the nature of the task, the datasets may inherently contain information about individuals (especially celebrities). Software and code for this paper is publicly available  at \url{https://github.com/clear-nus/edc}.
\paragraph{Human annotators.} The two annotators (1 male and 1 female) are recruited university students. The annotators are compensated fairly and given abundant and flexible time to complete the tasks. The collection protocol is determined exempt by our institution's IRB committee.
\paragraph{Potential Risks.} The use of current LLMs may incur risks such as hallucinations~\cite{xu2024hallucination} and privacy issues~\cite{yao2024llmprivacy}.

%% file: content/acks.tex
\section*{Acknowledgements}

This research is supported by the National Research Foundation Singapore and DSO National Laboratories under the AI Singapore Programme (AISG Award No: AISG2-RP-2020-016).

%% file: content/appendix.tex
\clearpage
\appendix

\section{Implementation Details}
\label{appendix: implementation_details}
\subsection{Models and Infrastructures Details}
\label{appendix: models_infrastructures_details}
We use two OpenAI models, GPT-3.5-turbo and GPT-4 (sizes currently unknown), and an open-source model, Mistral-7b (7 billion parameters). The training and inference of open-source models are done with a single machine with an AMD EPYC 7543P 32-Core Processor and 252GB of RAM, equipped with 4 NVIDIA RTX A6000 (48GB) GPUs. We accessed GPT-3.5-turbo and GPT-4 via the OpenAI API. Code for EDC is available at \url{https://github.com/clear-nus/edc}.

\subsection{Prompting-related hyperparameters}

We use few-shot prompting for all modules of EDC, we empirically choose 6-shot examples from the respective datasets. For the MCQ used in the Schema Canonicalization phase, we retrieve top-5 semantically similar relations from the schema as candidates. For refinement, the schema retriever retrieves top-10 most relevant relations from the schema as candidate relations. These hyperparameters are empirically chosen to balance performance and inference costs.

\subsection{Schema Retriever Training}
\label{appendix: schema_retriever_training}
We follow the original training methodology detailed in the original paper~\cite{wang2023e5mistral}, which involves utilizing pairs of text and their corresponding defined relations. For a given positive text-relation pair $(t^+,r^+)$, we employ an instruction template on $t^+$ to generate a new text $t^+_{inst} =$ ``Instruct: retrieve relations that are present in the given text $\textbackslash n$ Query: $\{t^+\}$''. 

We then finetune the embedding model to distinguish between the correct relation associated with a given text and other non-relevant relations using the InfoNCE loss,

\begin{align*}
\min \mathcal{L} = -\log \frac{\phi(t^+_{inst}, r^+)}{\phi(t^+_{inst}, r^+) + \sum_{n_i \in \mathbb{N}}\phi(t^+_{inst}, n_i)}
\end{align*}

Here, $\mathbb{N}$ denotes the set of negative samples, and $\phi$ represents the cosine similarity function. Please see the appendix for additional training details.

For training, we synthesized a dataset of text-relation pairs using the TEKGEN dataset~\citep{agarwal2020tekgen}, a large-scale text-triplets dataset created by aligning Wikidata triplets to Wikipedia text. The training dataset comprised 37,500 pairs, evenly divided between positive and negative samples. We adopted an online open-source implementation and hyperparameter configurations for training.

The performance of the fine-tuned schema retriever was assessed on the test splits of WebNLG, REBEL, and Wiki-NRE datasets. The recall@10 scores on these datasets were 0.823, 0.663, and 0.818, respectively, indicating the effectiveness of the retriever across different knowledge graph contexts.

\subsection{Details of Refinement Hint}
\label{appendix: refinement_hint_detail}
The refinement hint consists of candidate entities and candidate relations. This section details the obtainment of them and how they are used to improve the OIE performance. We will carry on using the example used in Section~\ref{sec: method}: ``\textit{Alan Shepard was born on Nov 18, 1923 and selected by NASA in 1959. He was a member of the Apollo 14 crew}'' and the triplets extracted by EDC in the first iteration are \textsf{['Alan Shepard’, ‘birthDate', ‘Nov 18, 1923’]}, \textsf{
['Alan Shepard’, 'mission', 'Apollo 14’]}.

\subsubsection{Obtaining Candidate Entities}

The candidate entities come from two sources:

\begin{itemize}
    \item Entities extracted by EDC from the previous iteration, i.e. \textsf{[`Alan Shepard’, `Nov 18, 1923’, `Apollo 14']}
    \item Entities extracted from the text by prompting the LLM to perform an entity extraction task, similar to the triplet extraction task.
    \begin{tcolorbox}[fontupper=\fon{cmss}, title=Entity Extraction Prompt]
\begin{footnotesize}
Given a piece of text, extract a list of entities from it.

Here are some examples:

Example 1:

Text: The 17068.8 millimeter long ALCO RS-3 has a diesel-electric transmission.

Entities: ['ALCO RS-3', 'Diesel-electric transmission', '17068.8 (millimetres)’]

...

Now please extract entities from the following text:
Alan Shepard was born on Nov 18, 1923 and selected by NASA in 1959. He was a member of the Apollo 14 crew.
\end{footnotesize}
\end{tcolorbox}
and the resultant entities are \textsf{[`Alan Shepard’, `Nov 18, 1923’, `NASA', `1959', `Apollo 14']}
\end{itemize}

The entities are then merged together as the candidate entities.

\subsubsection{Obtaining Candidate Relations}

The candidate relations also come from two sources:

\begin{itemize}
    \item Relations extracted by EDC from the previous iteration, i.e. \textsf{[`birthDate', `mission']}
    \item Relations extracted by the schema retriever, by calculating the relevance score between the input text and the relations in the schema. The top 5 retrieved relations in this case are \textsf{[`birthDate', , `selectedByNasa', `mission', `draftPick', `occupation']}.
\end{itemize}

The relations and their corresponding definitions are then merged together as the candidate relations. It is to be noted that, similar to other RAG-based methods, there is a chance that the retriever may retrieve irrelevant information. In this case, the relation definitions can come in handy as it provides more information for the LLMs to decide whether the relation is a valid one with respect to the text or not.

\subsubsection{Usage of Hint for Refined OIE}

The refinement hint is then included in the prompt appropriately to instruct the LLMs to consider (but is not limited to) the candidate entities and candidate relations:

\begin{tcolorbox}[fontupper=\fon{cmss}, title=Refined OIE Prompt, breakable]
\begin{footnotesize}
Given a piece of text, extract relational triplets in the form of [Subject, Relation, Object] from it.

Here are some examples:

Example 1:

Text: The 17068.8 millimeter long ALCO RS-3 has a diesel-electric transmission.

Entities: ['ALCO RS-3', 'Diesel-electric transmission', '17068.8 (millimetres)’]

Triplets: [['ALCO RS-3', 'powerType', 'Diesel-electric transmission'], ['ALCO RS-3', 'length', '17068.8 (millimetres)’]]

...

Now please extract triplets from the following text:
Alan Shepard was born on Nov 18, 1923 and selected by NASA in 1959. He was a member of the Apollo 14 crew.
Entities: [`Alan Shepard’, `Nov 18, 1923’, `NASA', `1959', `Apollo 14']

Here are some potential relations and their descriptions you may look out for during extraction: 
1. birthDate: The subject entity was born on the date specified by the object entity.

2. mission: The subject entity participated in the event or operation specified by the object entity.

3. selectedByNasa: The subject entity was selected by NASA in the year specified by the object entity.

...
\end{footnotesize}
\end{tcolorbox}

The extracted triplets by the refined OIE are:\textsf{['Alan Shepard’, ‘birthDate', ‘Nov 18, 1923’]}, \textsf{
['Alan Shepard’, 'mission', 'Apollo 14’]}, \textsf{['Alan Shepard', 'selectedByNasa', '1959']}. It successfully recovers the subtle and fine-grained relation \textsf{`selectedByNasa'} that would have been missed without using the hint. Also, the semantically rich descriptions help the LLM avoid excessively extracting noisy relations retrieved by the schema retriever.

We found it important to include the entities from both sources, i.e. extractions from the last round and discovered by a separate module (entity extraction or schema retriever). The significance of the schema retriever is already shown in the ablation study in Sec~\ref{target_alignment}.
 
\section{Annotation Instruction}

\begin{figure}
    \centering
\includegraphics[width=0.48\textwidth]
{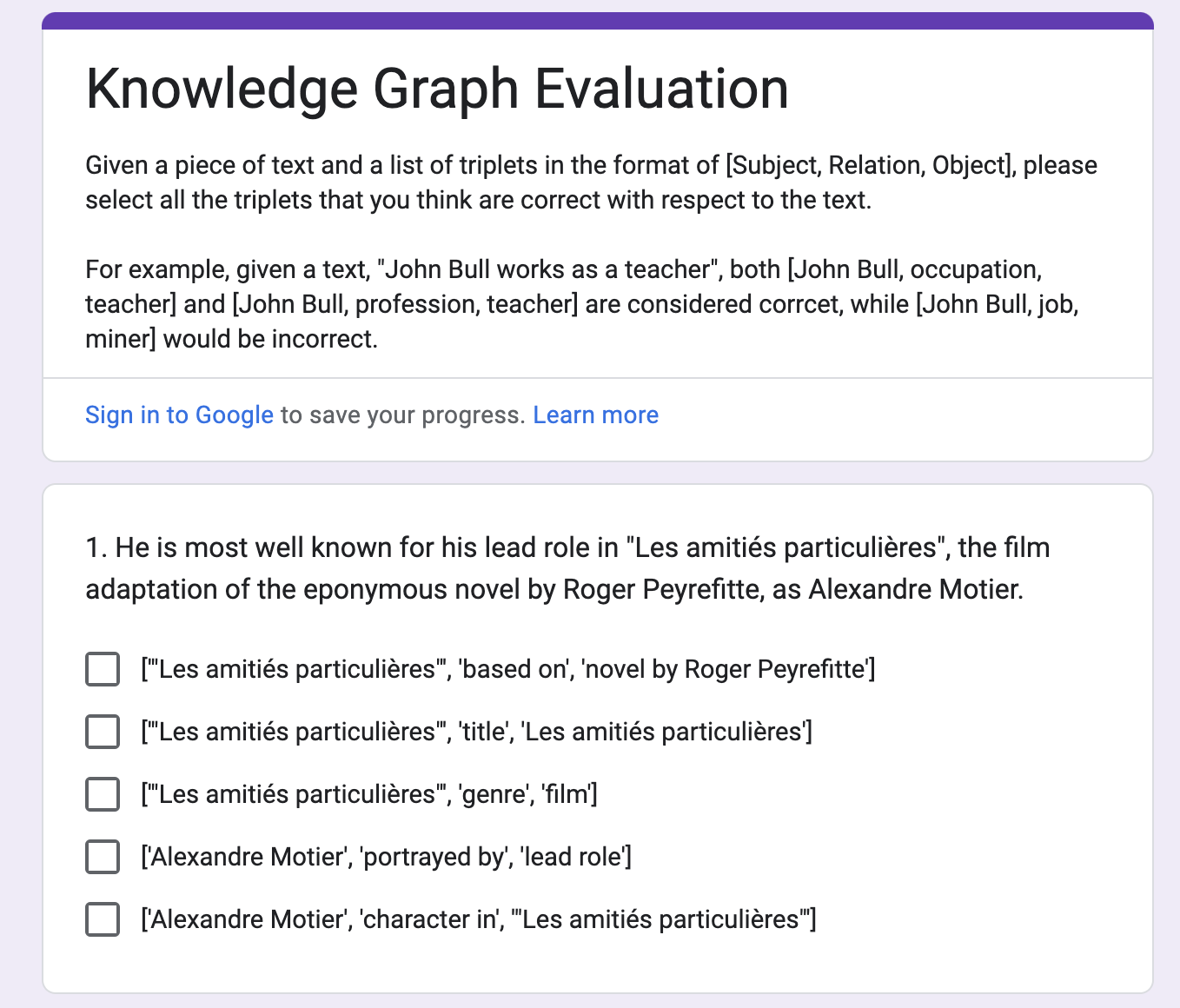}
    \caption{An example screenshot of the questionnaire including the instructions given to the annotators.}
    \label{fig: questionnaire}
\end{figure}

An example screenshot is provided in Figure~\ref{fig: questionnaire} to illustrate the format of questionnaires and instructions the annotators are given. The purpose of collection of the data was communicated to the annotators verbally.

\section{Detailed Results of Target Alignment}
\label{appendix: target_alignment_results}

\input{content/tbl_webnlg_full_results}
\input{content/tbl_rebel_full_results}
\input{content/tbl_wiki_full_results}
\input{content/tbl_more_refinement}

\input{content/tbl_no_last_round_extractions}

\input{content/tbl_number_of_extraction}

\subsection{Complete Results}

The complete results of \edc{} and \edcr{} on WebNLG, REBEL and Wiki-NRE are summarized in Table~\ref{tab: webnlg_complete}, Table~\ref{tab: rebel_complete} and Table~\ref{tab: wiki_complete} respectively. \edc{} performs better than or comparable to state-of-the-art baseline models in terms of all metrics (Precision, Recall, and F1) in all criteria (Partial, Strict, and Exact) and \edcr{} is able to consistently improve upon this in all aspects as well. These results more comprehensively demonstrate the performance of \edc{} and \edcr{}.

\subsection{Effect of More Refinement Iterations}

Table~\ref{tab: extra_refinement} shows the result of an extra iteration of refinement with \edc{} on all datasets. Although further refinement improves the results in a stable manner, we observe diminishing returns and hence, only report one iteration in the main results.

\input{content/tbl_novel_dataset_full_results}
\input{content/tbl_small_datasets_results}

\subsection{Ablation Study on Last-Round Extractions}
\label{appendix: ablation_last_round_extractions}
Table~\ref{tab: no_last_round_extractions} shows the result of ablating the relations and entities from the last round's extractions from the refinement hint. It shows the importance of performing the refinement in an iterative manner. Merging the two sources led to better coverage of the entities and relations in the text, resulting in better KGC.

\subsection{Discussion on KGC Dataset Annotations}

As stated in Section~\ref{result_subsection}, we observe that \edc{} is penalized by the scorer on Rebel and Wiki-NRE datasets due to incomplete annotations. This echoes the previous finding in~\citep{wadhwa2023revisiting, han2023isiesolvedbychatgpt} that LLMs can often extract correct results that are missing in the annotations, which results in overly pessimistic evaluations. As shown by Table~\ref{tab: number_of_triplets}, \edc{} tends to extract significantly more triplets compared to the reference annotations and the baseline GenIE. Furthermore, as shown from the manual evaluation in Table~\ref{tab: sc_results_vertical}, many of these triplets are indeed meaningful and correct with respect to the input text. Regardless, despite the automatic evaluation result on \edc{} being overly pessimistic, it still exceeds the baseline by a large margin and the actual performance may be even larger considering the difference in the number of triplets extracted.

\section{Experiments on a Novel Dataset}
\label{appendix: fictional_dataset_experiments}

Since the tested datasets were created  several years ago and the training set of the LLMs used are unknown, there is a risk the LLMs have already been trained on these  datasets. To address this concern, we create a novel small-scale dataset (50 entries) of fictional entities and information, e.g. ``\emph{Evergreen University was where Emily Johnson received her degree in Biology}'' and annotated them using the schema of Wiki-NRE. Table~\ref{tab: fictional_results} shows that EDC and EDC+R still obtain performance superior to the baseline GenIE model.

\section{Comparison against previous LLM-based approaches}
\label{appendix: small_datasets_results}
Although this is not the intended use scenario for EDC, we include these experimental results for a more comprehensive evaluation to compare against existing LLM-based methods. We conduct experiments on three datasets, CoNLL04 (4 relation types)~\cite{roth2004conll}, SciERC (7 relation types)~\cite{luan2018scierc} and our sub-sampled version of Wiki-NRE (45 relation types), which is the only dataset we use in our main experiments that can fit in the context window. To ensure comparison fairness, we use GPT-3.5-turbo for all the compared methods.

As shown in Table~\ref{tab: small_datasets_results}, when the relation number is small (CONLL and SciERC), EDC alone may not be superior to the baseline methods due to excluding the schema in the prompt. However, through refinement, EDC+R is able to achieve significantly better results. This may be attributed to the usage of the semantically rich relation descriptions in the refinement step. Specifically, it helps correct two types of errors that may occur during extraction: 1. the Definition step helps disambiguate homonyms, e.g., the word "follows" has two different meanings for "John follows Taoism" v.s. "John follows Mary". EDC changes the "follows" in "John follows Taoism" to "adheres to". 2. Using the relation definitions, we find the Refinement step corrects head-tail relation errors, e.g., for a relation "father", it is unclear if the subject or the object is the father, and the definition prevents inconsistent use. This error-correcting effect was not possible in previous methods.

When tested on Wiki-NRE, which has a moderately-sized schema, EDC already significantly outperforms the baseline methods, potentially due to the confusion of the LLMs when dealing with long context~\cite{liu2024lostinthemiddle}. Furthermore, we observe that ChatIE and CodeKGC may still output out-of-schema relation words although the entire schema is provided in the prompt, echoing the previous findings~\cite{wadhwa2023revisiting}.

\section{Combine EDC with other IE tools}
\label{appendix: edc_with_other_tools}
EDC can be integrated with other IE tools, including chunking, coreference, and entity de-duplication. This is beneficial in scenarios such as processing long documents that exceed the context window length of LLMs. We ran experiments on Re-DOCRED~\citep{tan2022redocred} by combining EDC with LingMess~\citep{lingmess}, a SOTA coreference resolution method and simple sentence-level chunking. We observed an increase of strict micro F1 score from 0.132  to 0.234, while directly prompting the LLMs only achieves 0.060.

We also explored the effect of entity deduplication in combination with EDC. We used CESI~\citep{vashishth2018cesi}, a SOTA post-hoc canonicalization method to deduplicate the entities in the resulting KGs from EDC. And we observe a slightly improved F1 score from 0.516 to 0.520 on the REBEL dataset under the `Partial' criteria.

\section{Combining OIE and Schema Definition}
\label{combining_oie_and_sd}
As an attempt to reduce the cost of EDC, we explored combining the OIE and Schema Definition steps. We previously separated these two steps because our preliminary experiments showed OIE to be more challenging and separating the two subtasks allowed us to use a more expensive model for OIE and a smaller, cheaper model for schema definition.
However, separate LLM calls increases latency of the pipeline (and cost if the same LLM is used). Also, making the LLMs output the definitions along with the extracted triples might improve consistency. In a further experiment combining EDC and Schema Definition on REBEL using GPT-3.5-turbo, we observed slight performance gains (0.516 to 0.518 under the `Partial' criteria) and token cost reduction ($\approx$ 3k to 2k tokens per example).

%% file: content/tbl_webnlg_full_results.tex
\begin{table*}\centering
\caption{Complete results of \edc{} and \edcr{} on WebNLG dataset against the baseline REGEN (Precision, Recall, F1 with `Partial', `Strict' and `Exact' criteria). EDC+R only performs 1 iteration of refinement. The best results are bolded.}\label{tab: webnlg_complete}
\footnotesize
\begin{tabular}{l|l|ccc|ccc|ccc}\hline\hline
& & &Partial & & &Strict & & &Exact & \\
Method&LLM for OIE &Precision &Recall &F1 &Precision &Recall &F1 &Precision &Recall &F1 \\\hline
&GPT-4 &\textbf{0.776} &\textbf{0.796} &\textbf{0.783} &\textbf{0.729} &\textbf{0.741} &\textbf{0.733} &\textbf{0.751} &\textbf{0.765}& \textbf{0.756} \\
EDC &GPT-3.5 &0.739 &0.760 &0.746 &0.684 &0.697 &0.688 &0.708 &0.722 &0.713 \\
&Mistral-7b &0.723 &0.739 &0.728 &0.668 &0.679 &0.672 &0.692 &0.703 &0.696 \\\hline
&GPT-4 &\textbf{0.814} & \textbf{0.831} &\textbf{0.820} &\textbf{0.782} &\textbf{0.794} &
\textbf{0.786} &\textbf{0.796} &\textbf{0.808} &\textbf{0.800} \\
EDC+R &GPT-3.5 &0.788 &0.806 &0.794 &0.749 &0.761 &0.753 &0.768 &0.781 &0.772 \\
&Mistral-7b &0.756 &0.775 &0.762 &0.716 &0.727 &0.720 &0.735 &0.747 &0.739 \\\hline
Baseline & REGEN &0.755 &0.788 &0.767 &0.713 &0.735 &0.720 &0.714 &0.738 &0.723 \\
\hline\hline
\end{tabular}
\end{table*}

%% file: content/tbl_rebel_full_results.tex
\begin{table*}\centering
\caption{Complete results of EDC and EDC+R on REBEL dataset against the baseline
REGEN (Precision, Recall, F1 with ‘Partial’, ‘Strict’, and ‘Exact’ criteria). EDC+R only
performs 1 iteration of refinement. The best results are bolded.}\label{tab: rebel_complete}
\footnotesize
\begin{tabular}{l|l|ccc|ccc|ccc}\hline\hline
& & &Partial & & &Strict & & &Exact & \\
Method&LLM for OIE &Precision &Recall &F1 &Precision &Recall &F1 &Precision &Recall &F1 \\\hline
&GPT-4 &\textbf{0.543} &\textbf{0.552} &\textbf{0.546} &\textbf{0.498} &\textbf{0.503} &\textbf{0.500} &\textbf{0.511} &\textbf{0.517} &\textbf{0.514} \\
EDC &GPT-3.5 &0.503 &0.512 &0.506 &0.448 &0.453 &0.449 &0.471 &0.476 &0.473 \\
&Mistral-7b &0.512 &0.523 &0.516 &0.450 &0.457 &0.453 &0.481 &0.488 &0.483 \\\hline
&GPT-4 &\textbf{0.599} &\textbf{0.606} &\textbf{0.601} &\textbf{0.557} &\textbf{0.561} &\textbf{0.559} &\textbf{0.572} &\textbf{0.576} &\textbf{0.574} \\
EDC+R &GPT-3.5 &0.556 &0.565 &0.559 &0.513 &0.519 &0.516 &0.527 &0.533 &0.529 \\
&Mistral-7b &0.525 &0.550 &0.531 &0.461 &0.462 &0.462 &0.506 &0.511 &0.505 \\\hline
Baseline &GENIE &0.381 &0.391 &0.385 &0.353 &0.361 &0.356 &0.362 &0.369 &0.364 \\\hline\hline
\end{tabular}
\end{table*}

%% file: content/tbl_wiki_full_results.tex
\begin{table*}\centering
\caption{Complete results of EDC and EDC+R on Wiki-NRE dataset against the baseline
REGEN (Precision, Recall, F1 with ‘Partial’, ‘Strict’, and ‘Exact’ criteria). EDC+R only
performs 1 iteration of refinement. The best results are bolded.}\label{tab: wiki_complete}
\footnotesize
\begin{tabular}{l|l|ccc|ccc|ccc}\hline\hline
& & &Partial & & &Strict & & &Exact & \\
Method &LLM for OIE &Precision &Recall &F1 &Precision &Recall &F1 &Precision &Recall &F1 \\\hline
 &GPT-4 &\textbf{0.682} &\textbf{0.686} &\textbf{0.683} &\textbf{0.675} &\textbf{0.679} &\textbf{0.677} &\textbf{0.676} &\textbf{0.680} &\textbf{0.678} \\
EDC&GPT-3.5 &0.645 &0.651 &0.647 &0.636 &0.640 &0.638 &0.638 &0.643 &0.640 \\
&Mistral-7b &0.644 &0.650 &0.647 &0.636 &0.640 &0.637 &0.637 &0.641 &0.639 \\\hline
 &GPT-4 &\textbf{0.712} & \textbf{0.715} &\textbf{0.713} & \textbf{0.708} & \textbf{0.710} &\textbf{0.709} &\textbf{0.708} & \textbf{0.711} &\textbf{0.709} \\
EDC+R&GPT-3.5 &0.691 &0.696 &0.693 &0.684 &0.688 &0.685 &0.685 &0.689 &0.687 \\
&Mistral-7b &0.661 &0.667 &0.663 &0.647 &0.652 &0.649 &0.656 &0.661 &0.658 \\\hline
Baseline &GENIE &0.482 &0.486 &0.484 &0.462 &0.464 &0.463 &0.477 &0.479 &0.478 \\
\hline\hline
\end{tabular}
\end{table*}

%% file: content/tbl_more_refinement.tex
\begin{table*}\centering
\caption{Results (F1 scores with all criteria) of further iterative refinement, the LLM used for OIE is GPT-3.5-turbo. EDC+2xR is \edc{} with 2 iterations of refinement.}\label{tab: extra_refinement}
\footnotesize
\begin{tabular}{l|ccc|ccc|ccc}\hline\hline
& &WebNLG & & &REBEL & & &Wiki-NRE & \\
Method &Partial &Strict &Exact &Partial &Strict &Exact &Partial &Strict &Exact \\ \hline
EDC+2xR &0.797 &0.761 &0.775 &0.564 &0.521 &0.535 &0.697 &0.689 &0.660 \\
EDC+R &0.794 &0.753 &0.772 &0.559 &0.516 &0.529 &0.693 &0.685 &0.657 \\
\hline
EDC &0.746 &0.688 &0.713 &0.506 &0.449 &0.473 &0.644 &0.634 &0.637 \\
\hline\hline
\end{tabular}
\end{table*}

%% file: content/tbl_no_last_round_extractions.tex
\begin{table*}\centering
\caption{Results (F1 scores with all criteria) of ablating the entities and relations extracted from the last round from the refinement hint, the LLM used for OIE is GPT-3.5-turbo. EDC+R-lastround is \edc{} with refinement but entities and relations extracted from the last round are removed from the refinement hint.}\label{tab: no_last_round_extractions}
\footnotesize
\begin{tabular}{l|ccc|ccc|ccc}\hline\hline
& &WebNLG & & &REBEL & & &Wiki-NRE & \\
Method &Partial &Strict &Exact &Partial &Strict &Exact &Partial &Strict &Exact \\ \hline
EDC+R &0.794 &0.753 &0.772 &0.559&0.516&0.529 &0.693 &0.685 &0.657 \\
EDC+R-lastround &0.748 &0.698 &0.720 &0.534 &0.485 &0.505 &0.634 &0.622 &0.625 \\
\hline
EDC &0.746 &0.688 &0.713 &0.506 &0.449 &0.473 &0.644 &0.634 &0.637 \\
\hline\hline
\end{tabular}
\end{table*}

%% file: content/tbl_number_of_extraction.tex
\begin{table*}[!t]\centering
\caption{The average number of triplets extracted per sentence on all three datasets. The baseline model for WebNLG is REGEN while the baseline for Rebel and Wiki-NRE is GENIE. Numbers in the brackets are the difference from the reference annotations.}\label{tab: number_of_triplets}
\footnotesize
\begin{tabular}{l|ccc}\hline\hline

 LLM for OIE &WebNLG &REBEL &Wiki-NRE \\\hline
GPT-4 &3.47(+0.04) &5.11(+1.11) &3.49(+0.63) \\
GPT-3.5 &3.44(+0.01) &5.01(+1.01) &3.49(+0.63) \\
Mistral7b &3.45+(0.02) &4.68(+0.68) &3.75(+0.89)\\\hline
Baseline &- &2.20(-1.80) &3.08(+0.22) \\\hline
Reference &3.43 &4.00 &2.86 \\\hline\hline
\end{tabular}
\end{table*}

%% file: content/tbl_novel_dataset_full_results.tex
\begin{table*}[!htp]\centering
\caption{Complete results of \edc{} and \edcr{} on the novel fictional dataset against the baseline GenIE (Precision, Recall, F1 with `Partial', `Strict' and `Exact' criteria). EDC+R only performs 1 iteration of refinement. The best results are bolded. The LLM used for OIE is GPT-3.5-turbo.}\label{tab: fictional_results}
\footnotesize
\begin{tabular}{l|ccc|ccc|ccc}\hline\hline
& &Partial & & &Strict & & &Exact & \\
Method&Precision &Recall &F1 &Precision &Recall &F1 &Precision &Recall &F1 \\\hline
EDC &0.731 &0.771 &0.751 &0.687 &0.704 &0.691 &0.702 &0.720 &0.707 \\
EDC+R &0.761 &0.782 &0.767 &0.733 &0.750 &0.738 &0.733 &0.750 &0.738 \\\hline
GenIE &0.521 &0.547 &0.530 &0.426 &0.443 &0.432 &0.467 &0.483 &0.472 \\
\hline\hline
\end{tabular}
\end{table*}

%% file: content/tbl_small_datasets_results.tex
\begin{table*}[!htp]\centering
\caption{Complete results of EDC, EDC+R on CONLL, SciERC and Wiki-NRE datasets against the previous LLM-based approaches, CodeKGC and ChatIE. The LLMs used here are GPT-3.5-turbo to ensure comparison fairness. The best results are bolded.}\label{tab: small_datasets_results}
\footnotesize
\begin{tabular}{l|l|ccc|ccc|ccc}\hline\hline
& & &Partial & & &Strict & & &Exact & \\
Dataset&Method &Precision &Recall &F1 &Precision &Recall &F1 &Precision &Recall &F1 \\
\hline
&EDC &0.536 &0.552 &0.543 &0.481 &0.491 &0.485 &0.503 &0.515 &0.509 \\
CONLL &EDC+R &\textbf{0.580} &\textbf{0.593} &\textbf{0.585} &\textbf{0.514} &\textbf{0.522} &\textbf{0.517} &\textbf{0.549} &\textbf{0.558} &\textbf{0.552} \\\cline{2-11}
&CodeKGC &0.542 &0.55 &0.545 &0.503 &0.506 &0.504 &0.542 &0.546 &0.543 \\
&ChatIE &0.463 &0.477 &0.468 &0.360 &0.366 &0.363 &0.418 &0.427 &0.421 \\
\hline
&EDC &0.389 &0.408 &0.395 &0.288 &0.301 &0.292 &0.352 &0.365 &0.357 \\
SciERC &EDC+R &\textbf{0.447} &\textbf{0.461} &\textbf{0.451} &\textbf{0.340} &\textbf{0.349} &\textbf{0.343} &\textbf{0.406} &\textbf{0.416} &\textbf{0.410} \\\cline{2-11}
&CodeKGC &0.389 &0.398 &0.392 &0.277 &0.283 &0.279 &0.346 &0.353 &0.349 \\
&ChatIE &0.351 &0.367 &0.357 &0.212 &0.221 &0.215 &0.294 &0.302 &0.297 \\
\hline
&EDC &0.645 &0.651 &0.647 &0.636 &0.640 &0.638 &0.638 &0.643 &0.640 \\
Wiki-NRE &EDC+R &\textbf{0.691} &\textbf{0.696} &\textbf{0.693} &\textbf{0.684} &\textbf{0.688} &\textbf{0.685} &\textbf{0.685} &\textbf{0.689} &\textbf{0.687} \\\cline{2-11}
&CodeKGC &0.611 &0.614 &0.612 &0.605 &0.607 &0.606 &0.607 &0.609 &0.608 \\
&ChatIE &0.569 &0.574 &0.571 &0.541 &0.545 &0.543 &0.553 &0.557 &0.555 \\
\hline\hline
\end{tabular}
\end{table*}